\documentclass[conference]{IEEEtran}
\usepackage{graphicx}
\usepackage{amsmath}
\usepackage{amssymb}
\usepackage{hyperref}
\usepackage{cite}
\usepackage{soul}
\usepackage{algpseudocode}
\usepackage{algorithm}
\usepackage{booktabs}
\usepackage{multirow} 
\usepackage{subfig}
\usepackage{graphicx}
\usepackage{makecell}
\usepackage{pifont}
\usepackage[table]{xcolor}
\usepackage{xspace}
\usepackage{float}

\usepackage[table,xcdraw]{xcolor}
\usepackage{siunitx}
\usepackage[most]{tcolorbox}
\definecolor{highlightcolor}{RGB}{255, 165, 0} 

\newcommand{\cmark}{\ding{51}} 
\newcommand{\xmark}{\ding{55}}

\newcommand{\defn}[1]{\textbf{\emph{#1}}}

\newcommand{\REQUIRE}{\Require}
\newcommand{\ENSURE}{\Ensure}
\newcommand{\ACP}{\emph{UCF}\xspace}

\title{Learning Robust Representations for Malicious Content Detection via Contrastive Sampling and Uncertainty Estimation}

\author{
    \IEEEauthorblockN{
        Elias Hossain\IEEEauthorrefmark{1}\IEEEauthorrefmark{3}, 
        Umesh Biswas\IEEEauthorrefmark{2}, 
        Charan Gudla\IEEEauthorrefmark{2},
        Sai Phani Parsa\IEEEauthorrefmark{3}
    }\\
     \IEEEauthorblockA{
        \IEEEauthorrefmark{1} College of Engineering and Computer Science,
        University of Central Florida, Orlando, USA,
         mdelias.hossain@ucf.edu\\
         \IEEEauthorrefmark{2} Department of Computer Science \& Engineering,
        Mississippi State University, MS State, USA\\
        ucb5@msstate.edu, gudla@cse.msstate.edu\\
          \IEEEauthorrefmark{3} Department of Computer Science,
        University of Nevada, Las Vegas, USA,
        sai.parsa@unlv.edu
    }\\
    \IEEEauthorrefmark{3} Corresponding author
}

\begin{document}

\maketitle

\begin{abstract}
We propose Uncertainty Contrastive Framework (\ACP), a Positive–Unlabeled (PU) representation learning framework that integrates uncertainty-aware contrastive loss, adaptive temperature scaling, and a self-attention–guided LSTM encoder to improve classification under noisy and imbalanced conditions. \ACP dynamically adjusts contrastive weighting based on sample confidence, stabilizing training using positive anchors, and adapts temperature parameters to batch-level variability. Applied to malicious content classifications, \ACP-generated embeddings enabled multiple traditional classifiers to achieve over $93.38\%$ accuracy, precision above $0.93$, and near-perfect recall, with minimal false negatives and competitive ROC–AUC scores. Visual analyses confirmed clear separation between positive and unlabeled instances, highlighting the framework’s ability to produce calibrated, discriminative embeddings. These results position \ACP as a robust and scalable solution for PU learning in high-stakes domains such as cybersecurity and biomedical text mining.

\end{abstract}


\section{Introduction}

\noindent As digital ecosystems grow increasingly complex and interconnected, detecting malicious content, such as adversarial malware \cite{rosenberg2021adversarial}, fraudulent URLs \cite{ruchay2023imbalanced}, and coordinated misinformation, has emerged as a critical challenge. For traditional Machine Learning (ML) algorithms \cite{kamath2018comparative}, it is challenging to detect malicious content using only a few key ingredients (i.e., privacy, scalability, and labeling costs). Consequently, \emph{PU} \cite{li2009positive} \cite{li2022positive} \cite{kiryo2017positive} learning has become a promising paradigm, allowing classifiers to learn from a limited set of positively labeled data while treating the remainder as unlabeled \cite{jaskie2022positive}. 

PU learning is a variant of binary classification \cite{hsieh2019classification} where the training data contains only positively labeled and unlabeled examples, with no explicitly labeled negatives \cite{bekker2020learning}. This setup reflects many real-world scenarios where negative labels are unavailable or unreliable, such as in personalized advertising, electronic medical records, and knowledge base completion \cite{bekker2020learning}. Unlike one-class classification \cite{seliya2021literature} \cite{perera2019learning} or traditional semi-supervised learning \cite{zhu2005semi}, PU learning explicitly leverages unlabeled data, assuming that it may contain both positive and negative instances. As such, it occupies a critical space in the broader field of weakly supervised learning \cite{zhou2018brief} and has gained growing attention for its practical significance and theoretical challenges, including how to estimate class priors, formalize learning objectives, and evaluate models without full label information \cite{elkan2008learning,du2014class}. 

Despite recent progress, most PU learning approaches make strong assumptions about label quality and class separability, which do not hold in practice. Furthermore, standard deep learning approaches tend to underperform in weakly supervised scenarios, especially when exposed to label noise and structural variation in the anomaly distribution. 

Meanwhile, \defn{contrastive learning} has shown promise in self-supervised representation learning by maximizing similarity between semantically related instances \cite{vinay2022fraud,chen2020simple,saunshi2019theoretical,acharya2022positive}. However, standard contrastive methods are not directly compatible with PU settings, as the absence of reliable positive and negative labels introduces significant uncertainty in supervision.

To address these limitations, \ACP, a novel framework integrating \defn{contrastive representation learning} with \defn{uncertainty-aware PU modeling}, is proposed. The contributions are as follows:

\begin{itemize}
    \item An \defn{uncertainty-weighted contrastive loss} is introduced, dynamically adjusting the importance of sample pairs based on their confidence scores, thereby mitigating the effects of label ambiguity.
    \item An \defn{adaptive temperature scaling} strategy is designed to modulate the contrastive learning signal over time, ensuring training stability across noisy and imbalanced data.
    \item A \defn{self-attention-based encoder} is incorporated to enhance the model’s ability to capture complex temporal and structural patterns, further improving representation robustness.
\end{itemize}

\textbf{Outline of the paper.} The paper begins with a review of previous work in PU learning and its applications in Section~\ref{sec:related_work}. In Section~\ref{uncertainty_contrastive_framework}, \ACP is presented, including the formal problem setup and the proposed \ACP methodology. The experimental setup, encompassing dataset construction, preprocessing, and evaluation protocols, is described in Section~\ref{experimental_setup}. Details of data acquisition and cleaning, model evaluation techniques, and implementation settings are also provided. Section~\ref{sec:results} reports the experimental findings and diagnostic insights, with an emphasis on the robustness and discriminative capability of the learned embeddings. The paper concludes in Section~\ref{conclusion_futureWork}, where potential directions for future research are outlined.

\section{Related Work}
\label{sec:related_work}

Several studies have integrated PU learning with deep learning to improve intrusion and malware detection under imbalanced conditions. Fan et al.~\cite{fan2023self} introduced Self-paced and Reweighting PU learning (SRPU), which selects confident samples and adaptively reweights losses, achieving an F1 score of 0.9451 and a Matthews Correlation Coefficient (MCC) of 0.9363 on three real-world datasets. Go et al.~\cite{go2020visualization} presented a visualization-based malware classification technique using ResNeXt on grayscale malware binaries, reaching 98.86\% accuracy on the Malimg dataset. Zhang et al.~\cite{zhang2019network} developed a deep hierarchical network combining Convolutional Neural Networks (CNNs) and Long Short-Term Memory (LSTM) layers to capture spatial-temporal features from raw network flows, reporting up to 99.98\% recall on the CICIDS2017 and CTU datasets.

Similarly, Zeng et al.~\cite{zeng2019deep} proposed Deep-Full-Range (DFR), which combines 1D CNN, LSTM, and Stacked Autoencoder (SAE) architectures for encrypted traffic classification and intrusion detection. DFR achieved an F1 score of 0.9987 and 99.41\% accuracy on ISCX datasets. Nayyar et al.~\cite{nayyar2020recurrent} implemented a 14-layer Recurrent Neural Network (RNN) model with LSTM units for real-time intrusion detection, attaining 96.7\% accuracy on the CAIDA DDoS 2007 dataset with an ultra-low prediction latency of 7 $\mu$s.

PU learning has also shown promise in web security applications. Wang et al.~\cite{wang2024pu} introduced a PU-based framework for detecting actual exploitation in cross-site scripting (XSS) attacks by analyzing server responses with handcrafted features and a Gradient Boosting Decision Tree (GBDT) ensemble. The model achieved 86\% precision and 94\% recall across five vulnerable web applications. Zhang et al.~\cite{zhang2017poster} proposed a PU-based malicious URL detection method, incorporating a two-stage negative sample selection strategy and cost-sensitive logistic regression. Their system, tested on production traffic from Ant Financial, achieved up to 94.2\% accuracy and uncovered new attack patterns missed by traditional detection systems.

In the medical domain, PU learning has been successfully applied to both imaging and genomic data. Zhang et al.~\cite{zhang2024semi} proposed Hölder Divergence-based Positive and Negative Learning (HD-PAN), a semi-supervised framework for disease classification using medical image data. Unlike traditional Generative Adversarial Networks (GANs), HD-PAN employs a classifier guided by Hölder divergence instead of Kullback–Leibler divergence, achieving classification accuracies between 91.02\% and 98.87\% on datasets such as BreastMNIST and PneumoniaMNIST. Yang et al.~\cite{ren2014positive} introduced Positive-Unlabeled Learning for Disease Gene Identification (PUDI), which partitions unlabeled genes into confidence-based subsets and applies a multi-level Support Vector Machine (SVM) using biological features such as Gene Ontology annotations and protein–protein interaction networks. PUDI achieved an F-measure of 76.5\%, outperforming earlier methods like ProDiGe and K-Nearest Neighbors (KNN) in general and specific disease gene prediction tasks.

These studies demonstrate the adaptability of PU learning across domains, particularly where labeled data are limited and the cost of false negatives is high.

\section{Uncertainty Contrastive Framework (UCF)}

\label{uncertainty_contrastive_framework}

In the PU learning context, contrastive learning faces two primary challenges: uncertainty in labels and variability in anomaly distributions. Conventional methods assume label accuracy, which often degrades representation quality when supervision is noisy. The proposed \ACP framework addresses this by integrating uncertainty estimation into the contrastive learning objective, using uncertainty scores to reweight pair contributions. This mechanism prioritizes confident positive samples and down-weights ambiguous or noisy instances, resulting in representations that remain both robust and discriminative. By uniting contrastive learning with uncertainty modeling, \ACP enables more reliable and interpretable feature extraction in PU scenarios.

\subsection{Problem Setup and Notations}

Consider $\Omega = \mathbb{R}^d$ as the domain of input, consisting of feature vectors in $d$ dimensions, and let the target space be $\Lambda = \{-1, +1\}$, where $+1$ indicates anomalous data and $-1$ signifies normal data points. Suppose we have a training dataset $\Xi$ with $T$ samples. The initial $r$ samples are labeled as anomalies, denoted by $\Xi_a = \{(s_1, t_1), (s_2, t_2), \ldots, (s_r, t_r)\}$, where each $s_j \in \Omega$ and $t_j = +1$. The remaining $T - r$ samples are unlabeled and represented by $\Xi_u = \{s_{r+1}, s_{r+2}, \ldots, s_T\}$. Although all labeled samples are marked as anomalous (i.e., $t = +1$), their inherent features can vary widely, indicating different types of anomalies. The primary objective is to create a classification function $g : \Omega \rightarrow \Lambda$ that can effectively distinguish between anomalous and normal samples, even when the anomalies differ in structure or behavior. The model should be capable of generalizing to new inputs, ensuring reliable anomaly detection in future applications.

\begin{table}[t]
\centering
\caption{Summary of symbols used in the contrastive PU formulation.}
\label{tab:contrastive-symbols}
\begin{tabular}{ll}
\toprule
\textbf{Symbol} & \textbf{Description} \\
\midrule
$\Omega$ & Input feature space, $\mathbb{R}^d$ \\
$\Lambda$ & Label space, $\{-1, +1\}$ \\
$D^1$ & Set of labeled positive samples \\
$D^U$ & Set of unlabeled samples \\
$S$ & Training batch from $D^1 \cup D^U$ \\
$S^a$ & Auxiliary batch sampled from $D^1$ \\
$z_i$ & Embedding of sample $x_i$ \\
$A(x_i)$ & Candidate set for $x_i$ \\
$B_1(x_i)$ & Positive set for $x_i$ \\
$B_0(x_i)$ & Negative set for $x_i$ \\
$\tau$ & Adaptive temperature parameter \\
$w(x_i)$ & Uncertainty weight for $x_i$ \\
$\mathcal{L}^{PU}$ & Uncertainty-aware PU contrastive loss \\
\bottomrule
\end{tabular}
\end{table}

\subsection{Proposed Approach}

This section introduces the implemented \ACP framework, which offers an innovative strategy for representation learning in PU settings. As illustrated in Algorithm~\ref{alg:adaptive_conpu}, our approach incorporates adaptive temperature scaling, a self-attention mechanism within the encoder, and an uncertainty-aware contrastive loss to improve both robustness and generalization in scenarios with limited labels. The methodology unfolds in five key stages: representation extraction, adaptive temperature computation, formation of contrastive pairs, uncertainty-aware loss design, and encoder training. We now describe each of these components in detail.

\begin{figure*}[t]
\includegraphics[width=\textwidth]{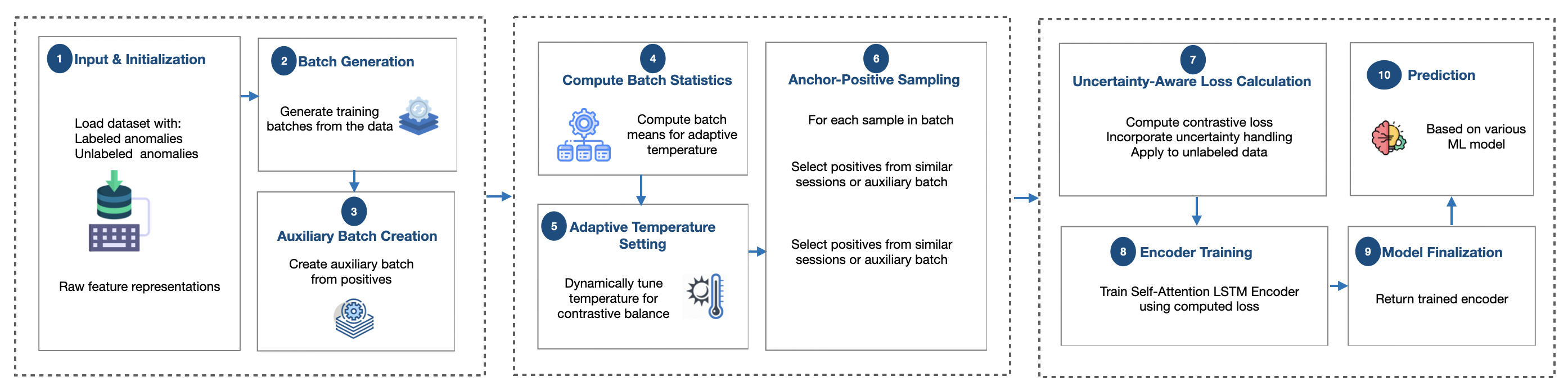}
\vspace{-15pt}\caption{The proposed Uncertainty Contrastive Framework (UCF) for malicious content detection. The framework integrates preprocessing and sampling, adaptive temperature scaling, uncertainty-weighted contrastive representation learning, and a self-attention LSTM encoder to generate discriminative embeddings used for final classification.}
\label{algo-range-visualise}
\end{figure*}

\vspace{5pt}
\subsubsection{Representation Extraction and Batch Formation}
\label{subsubsec:representation-extraction}

For each session in the dataset \( D = D^1 \cup D^U \), we start by creating initial representations.  \( D^1 \) indicates the labeled (positive) section, and \( D^U \) indicates the unlabeled portion.  We create training batches \( S = \{x_i\}_{i=1}^{R} \) from this complete dataset, where $R$ indicates the number of samples in each training batch.  We generate an auxiliary batch \( S^a = \{x_i^1\}_{i=1}^{M} \) for every training batch, which is solely derived from \( D^1 \).  During contrastive learning, these auxiliary samples serve as anchors, assisting in maintaining training stability. In addition to preparing the model for contrastive learning, this representation extraction step also sets the foundation for the model's subsequent adaptive management of data variability.

\vspace{5pt}

\subsubsection{Adaptive Temperature Scaling}
\label{subsubsec:adaptive-temperature-scaling}

Conventional contrastive methods control the sharpness of the similarity distribution by using a constant temperature parameter (\( \tau \)).  On the other hand, we suggest \textbf{adaptive temperature scaling}, which dynamically adjusts \( \tau \) according to the representation space as follows: 

\begin{equation}
\label{eq:adaptive-tau}
\tau = \frac{\sigma(\mathbf{v}_D)}{\log(1 + \text{epoch})}
\end{equation}

where the mean representation of the current training batch is represented by \( \mathbf{v}_D = \frac{1}{R} \sum_{i=1}^{R} z_i \) and the standard deviation across its dimensions is indicated by \( \sigma(\mathbf{v}_D) \).

 We establish a variance-adjusted movement direction in the representation space to better capture batch-level dynamics as follows:

\begin{equation}
\mathbf{v}_0 = \frac{1}{\tau^0} \left[ \mathbf{v}_D - \tau^1 \mathbf{v}_1 \right], \quad \text{where } \mathbf{v}_1 = \frac{1}{M} \sum_{i=1}^{M} z_i^1
\end{equation}

Here, the mean embedding from the auxiliary batch of labeled positives \( D^1 \) is indicated by \( \mathbf{v}_1 \). The directional shift in the embedding space between the full batch and the positives-only batch is represented by the vector \( \mathbf{v}_0 \), which is modulated by two scaling constants: \( \tau^0 \) and \( \tau^1 \). More specifically, \( \tau^0 \) determines the directional adjustment's overall scale, whereas \( \tau^1 \) affects the function of the positive anchor representations.

During training, this dynamic scaling technique ensures that contrastive gradients remain both stable and adaptable, especially in the presence of noisy or diverse data distributions. We create positive and negative pairings after defining these parameters, which are essential for uncertainty-aware contrastive learning.

\vspace{5pt}

\subsubsection{Construction of Positive and Negative Sets}
\label{subsubsec:positive-negative-construction}

Following the acquisition of the representations, each sample's candidate positive and negative sets need to be formed \( x_i \in S \).  The candidate set is defined as follows: \( A(x_i) = (S \cup S^a) \setminus \{x_i\} \). It is then split into two subsets using a predictive labeling process. We provide a summary of the fundamental notations in Table~\ref{tab:contrastive-symbols} following to help the reader comprehend the mathematical symbols used during this process:

This symbolic paradigm facilitates the creation of contrastive sets that balance labeled and unlabeled data while taking uncertainty into account.  It guarantees that when uncertainty is present, the model highlights significant differences and adjusts the loss function appropriately. The resulting uncertainty-aware PU contrastive loss is defined as follows:

\begin{equation}
\resizebox{0.95\linewidth}{!}{
$\displaystyle
\mathcal{L}^{PU} 
= \frac{1}{R} \sum_{i=1}^R 
  \Bigl[
     I(x_i) \sum_{x_p \in B_1(x_i)} \ell(z_i, z_p) 
     + 
     (1 - I(x_i)) \sum_{x_n \in B_0(x_i)} \ell(z_i, z_n)
  \Bigr]
$
}
\end{equation}

For robust representation learning to occur in positive-unlabeled environments, the composition of these sample groups is essential and directly affects the loss function's performance.  For robust representation learning in positive-unlabeled situations, the ordering of these subsets is essential for estimating the efficacy of the loss function.


\vspace{5pt}
\subsubsection{Uncertainty-Aware Contrastive PU Loss}
\label{subsubsec:uncertainty-aware-loss}

In PU learning settings, the distinction between positive and unlabeled cases is often unclear, leading to inaccurate pairwise comparisons.  To get around this, we use a contrastive PU loss that is aware of uncertainty and adjusts each pair's contribution based on how close it is to the decision border.  This formulation defines the uncertainty weight \( w(x_i) \) as \( w(x_i) = 1 - \max(p(x_i)) \), where \( p(x_i) \) represents the expected probability distribution over the classes.  The model can concentrate more on ambiguous cases close to the decision border because of this weighting mechanism, which gives samples with lower confidence a higher priority.

\begin{equation}
\resizebox{0.95\linewidth}{!}{$
\displaystyle
\begin{aligned}
\mathcal{L}^{PU} 
&= \frac{1}{R} \sum_{i=1}^{R} \Biggl[
   \mathbb{I}(x_i) \frac{1}{\lvert B_1(x_i) \rvert} \sum_{x_p \in B_1(x_i)} \ell(z_i, z_p) \\[1mm]
&\quad +\, (1 - \mathbb{I}(x_i)) \frac{1}{\lvert B_0(x_i) \rvert} \sum_{x_p \in B_0(x_i)} \ell(z_i, z_p)
\Biggr] \cdot w(x_i)
\end{aligned}
$}
\end{equation}

In this case, the uncertainty weight highlights samples close to the classification border, and \(\ell(z_i, z_p) \) denotes the contrastive loss calculated between the representation of \(x_i \) and \(x_p \). This encourages the encoder to concentrate more on ambiguous cases, which enhances calibration and increases the model's reliability in noisy PU scenarios found in the real world.

Having defined the objective function, we now turn to the architecture and training procedure used to optimize it.


\vspace{5pt}
\subsubsection{Encoder Architecture and Training}
\label{subsubsec:encoder-training-architecture}

The LSTM encoder is based on self-attention and processes the input sessions. By including a self-attention layer, the model's ability to identify contextual patterns and long-term dependencies in sequential data is enhanced. Specifically tailored for contrastive learning, our attention-guided architecture generates more informative embeddings. After training the encoder with the batch-wise uncertainty-aware loss \(\mathcal{L}^{PU} \), the encoder weights are iteratively updated to improve the robustness and quality of the learned representations. Upon completion of the \ACP training procedure, the result is a well-calibrated encoder with improved accuracy in distinguishing between positive and unlabeled data.

\begin{algorithm}
\small
\caption{Adaptive Contrastive PU Learning}
\label{alg:adaptive_conpu}
\begin{algorithmic}[1]
\REQUIRE $D = D^1 \cup D^U$, adaptive $\tau$, $R$, $M$, and an untrained encoder with self-attention.
\ENSURE Trained encoder with improved feature extraction and uncertainty-aware contrastive learning.

\State Generate raw representations of all sessions in $D$.
\State Generate training batches from $D$.
\For{each training batch $S = \{x_i\}_{i=1}^{R}$}
    \State Create auxiliary batch $S^a = \{x_i\}_{i=1}^{M}$ from $D^1$.
    \State Compute: $\mathbf{v}_D = \frac{1}{R} \sum_{i=1}^{R} \mathbf{z}_i$, \quad $\mathbf{v}_1 = \frac{1}{M} \sum_{i=1}^{M} \mathbf{z}_i$
    \State Adaptive temperature: $\tau = \frac{\sigma(\mathbf{v}_D)}{\log(1 + \text{epoch})}$
    \State Compute: $\mathbf{v}_0 = \frac{1}{\tau^0} [ \mathbf{v}_D - \tau^1 \mathbf{v}_1 ]$
    \For{each session $x_i \in S^0$}
        \State $A(x_i) = (S \cup S^a) \setminus \{x_i\}$
        \State Initialize $B_0(x_i) = \emptyset$
        \For{each $x_j \in A(x_i)$}
            \If{$\mathbb{I}(x_j) == 0$}
                \State $B_0(x_i) = B_0(x_i) \cup x_j$
            \EndIf
        \EndFor
        \State $B_1(x_i) = A(x_i) - B_0(x_i)$
    \EndFor

    \State \textbf{Uncertainty-Aware Contrastive PU Loss:}
    \Statex
    \begin{equation*}
    \begin{aligned}
    \mathcal{L}^{PU} = \frac{1}{R} \sum_{i=1}^{R} \Biggl[
    &\mathbb{I}(x_i) \frac{1}{|B_1(x_i)|} \sum_{x_p \in B_1(x_i)} \ell(z_i, z_p) \\
    &+ (1 - \mathbb{I}(x_i)) \frac{1}{|B_0(x_i)|} \sum_{x_p \in B_0(x_i)} \ell(z_i, z_p)
    \Biggr] \cdot w(x_i)
    \end{aligned}
    \end{equation*}

    \State Train the Self-Attention LSTM Encoder using this batch loss.
\EndFor
\State \Return trained encoder.
\end{algorithmic}
\end{algorithm}

\section{Experimental Setup}

\label{experimental_setup}

\subsection{Data Acquisition and Cleaning}

This study utilized a synthetically generated dataset produced through a prompt-based engineering pipeline leveraging a large language model (LLM). A total of 15,000 samples were created, each comprising ten numerical features, a session identifier, and an associated label. These synthetic instances were crafted to simulate realistic patterns commonly encountered in binary classification tasks with partial supervision. To emulate a PU learning setting, 1,000 samples were explicitly labeled as positive, while the remaining 14,000 instances were treated as unlabeled. This class distribution reflects practical scenarios where positive labels are limited and the majority of data lacks annotation.

Standard data preprocessing techniques were applied prior to model development. These included the removal of duplicate records, validation of numerical feature integrity, and verification that neither session identifiers nor label metadata contaminated the feature space. All features were normalized to a uniform scale to ensure consistent convergence across classifiers. The final dataset was structured to support PU learning pipelines, enabling the coexistence of labeled positives and unlabeled instances within an uncertainty-aware modeling framework.

\subsection{Model Evaluation Techniques}

To ensure methodological rigor, a 5-fold cross-validation strategy was employed for evaluating classifier performance across all experimental conditions. Through this approach, sampling bias was mitigated and a reliable estimation of generalization capability was obtained. For each fold, models were trained and evaluated using the embeddings produced by the proposed \ACP framework. In addition to standard evaluation metrics such as accuracy, precision, recall, and F1 score, the area under the receiver operating characteristic curve (ROC-AUC) was incorporated to assess each model's discriminative capacity across varying decision thresholds.  

Building upon this evaluation framework, a deeper examination of classification behavior was conducted through the generation of confusion matrices for each model and fold. The predictions were decomposed into \textit{true positives} (TP), \textit{false positives} (FP), \textit{false negatives} (FN), and \textit{true negatives} (TN) by these matrices. False positives represent negative samples incorrectly classified as positive, false negatives represent positive samples incorrectly classified as negative, and true negatives represent negative samples correctly predicted. True positives represent positive samples correctly predicted. This analysis allowed for a more detailed evaluation of model reliability by highlighting the distribution of classification errors in addition to the overall accuracy.  

To complement the quantitative analysis, a qualitative assessment of the learned feature space was carried out using t-distributed Stochastic Neighbor Embedding (t-SNE) on the validation set embeddings. After the final training stage, this visualization was used to evaluate the separability of positive and negative classes in the reduced two-dimensional space, providing an interpretable complement to the numerical performance metrics.

\subsection{Implementation Details}

To provide clarity on the experimental reproducibility and computational environment, the implementation setup is outlined as follows. All experiments were implemented in Python using the \texttt{scikit-learn} and \texttt{PyTorch} libraries. Model training and inference were conducted on a Microsoft Azure Standard NC24ads~A100~v4 virtual machine equipped with an NVIDIA~A100 GPU (80~GB HBM2e memory), 24~vCPUs, and 220~GiB of system RAM, running Ubuntu~24.04~LTS. For optimization, the Adam optimizer was used with a learning rate of $1\times 10^{-4}$ and a batch size of 32 unless otherwise specified. Random seeds were fixed across all runs to ensure reproducibility.

\section{Results}
\label{sec:results}

\subsection{Training Dynamics and Embedding Evaluation}

\begin{figure}[H]
    \centering
    \includegraphics[width=\linewidth]{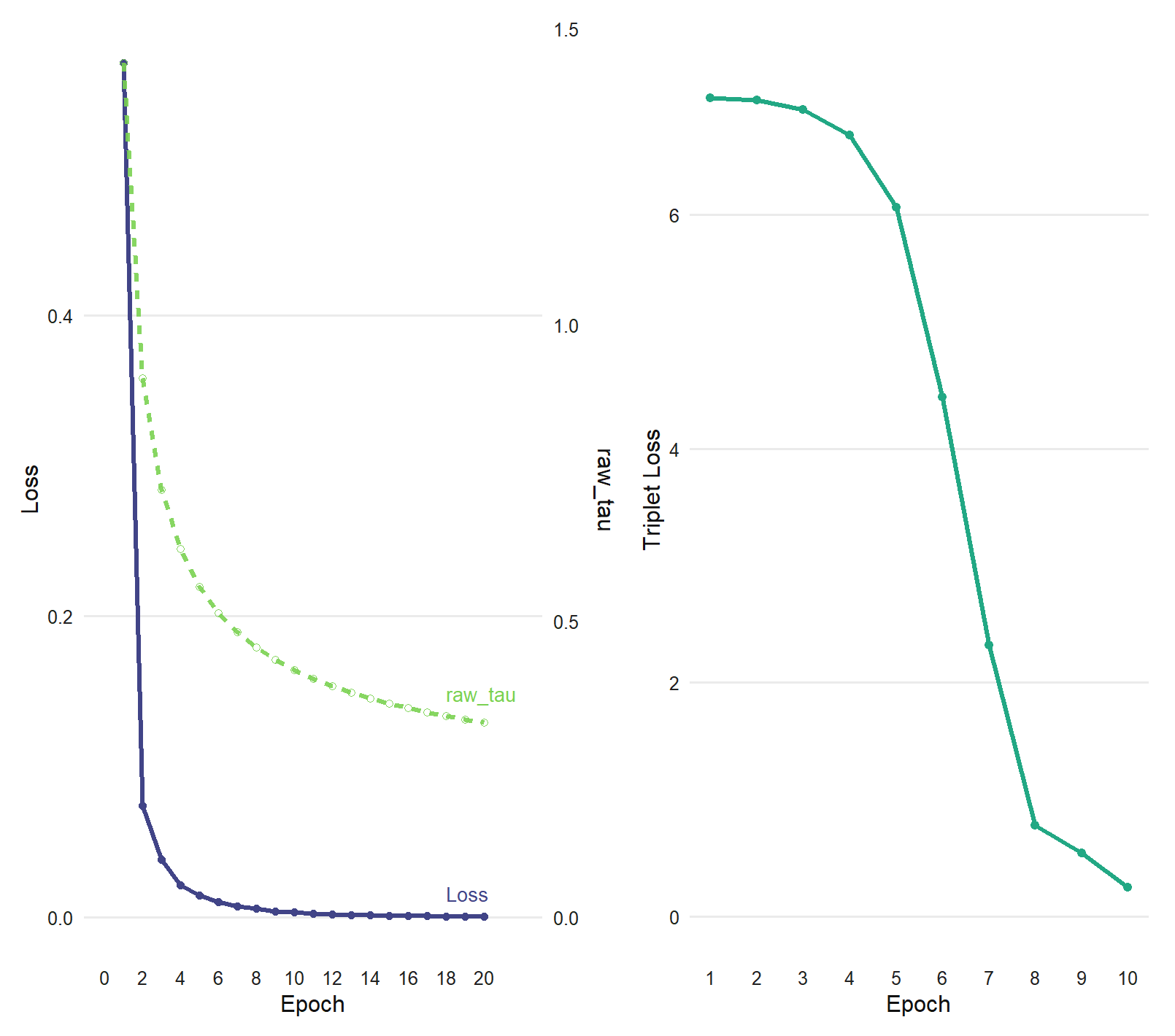}
    \caption{Training loss curves for Stage~1 (ConPU loss and raw $\tau$) and Stage~2 (pseudo-negative triplet loss).}
    \label{fig:stage1_stage2}
\end{figure}

The model was trained in two consecutive phases, each designed to address a distinct optimization objective, as illustrated in Fig.~\ref{fig:stage1_stage2}.  

During \textbf{Stage~1}, the ConPU loss on $D_1$ relative to the unlabeled training set decreased sharply from approximately $0.52$ at epoch~1 to about $0.07$ by epoch~2, and further dropped below $10^{-2}$ around epoch~10. This rapid reduction indicates that stable separation between positive and unlabeled instances was achieved early in training. Concurrently, the raw $\tau$ parameter followed a smooth downward trend, confirming consistent convergence of the pairwise ranking margin throughout this phase.  

In \textbf{Stage~2}, the pseudo-negative triplet loss remained near $7.0$ during the initial five epochs, reflecting early difficulties in enforcing large inter-class margins. From approximately epoch~6 onward, the loss declined markedly, reaching below $1.0$ by the final epoch. This progression demonstrates the model’s improved ability to distinctly position positive samples apart from pseudo-negative embeddings, thereby producing a more discriminative latent space.

\subsection{Embedding Space Visualization}

\begin{figure}[t]
  \centering
  \includegraphics[width=\linewidth]{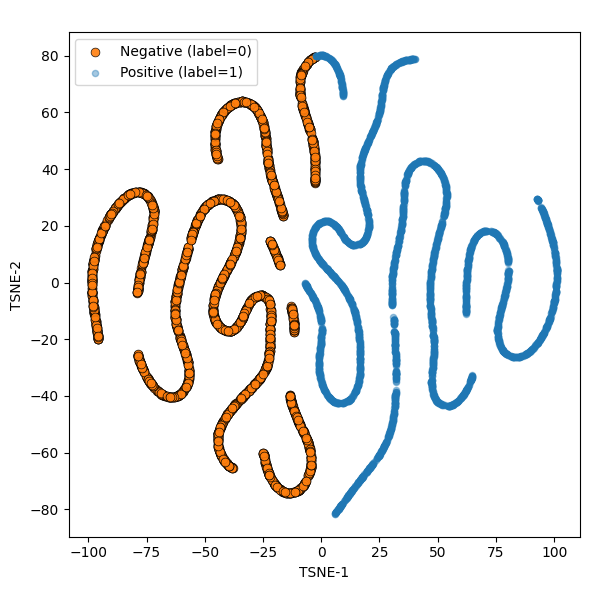}
  \caption{t-SNE of unlabeled\_train embeddings after Stage~2.}
  \label{fig:tsne_train}
\end{figure}

To qualitatively evaluate the ability of the learned representations to distinguish between classes, t-SNE was applied to map the embeddings obtained after Stage~2 into a two-dimensional space. As depicted in Fig.~\ref{fig:tsne_train}, the \textit{unlabeled\_train} set displays two distinct clusters corresponding to positive and negative instances. The tight clustering within each class and the lack of significant overlap indicate that class-specific structures in the latent space were effectively captured by the encoder, even with the inherent uncertainty in PU learning.  

\begin{figure}[t]
  \centering
  \includegraphics[width=\linewidth]{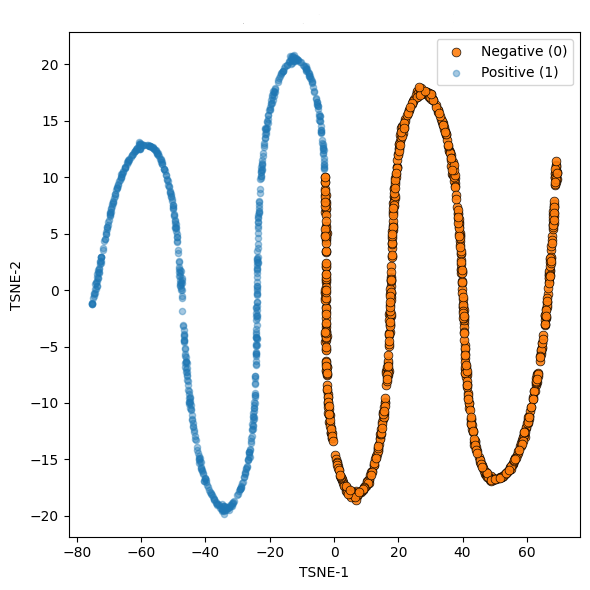}
  \caption{t-SNE of validation embeddings after Stage~2.}
  \label{fig:tsne_val}
\end{figure}

A similar pattern is evident in the held-out validation set (Fig.~\ref{fig:tsne_val}), where positive and negative samples remain distinctly separated, suggesting that the learned feature space generalizes beyond the training distribution. Notably, both splits exhibit smooth, continuous cluster boundaries, implying that the representation space is not only separable but also preserves meaningful manifold geometry.  

These visualizations are consistent with the quantitative results reported earlier: high classification accuracy, minimal false negatives, and strong ROC–AUC scores. The clear separability observed in the t-SNE plots further confirms that the \ACP framework produces calibrated, discriminative embeddings that maintain their structure across unseen data, a crucial property for robust deployment in real-world PU learning scenarios.

\subsection{Freezing LSTM Layers and Classification Performance}

In this investigation, the LSTM encoder was maintained at \emph{frozen}, and its embeddings were employed as fixed features for a variety of conventional classifiers. The quality of the learned representations is thereby isolated from any supplementary fine-tuning effects.

According to Table~\ref{tab:exp1_frozen_cls}, the F1 scores of $\approx 0.9657$ were attained by \textit{Logistic Regression}, \textit{SVM}, and \textit{Gradient Boosting}, which achieved the highest accuracy ($93.38\%$) and near-perfect recall. Similar results were observed with \textit{K-Nearest Neighbors} ($93.25\%$ accuracy), indicating that the embedding space is well-organized for both distance-based and linear classification. Although \textit{Random Forest} attained a slightly lower accuracy ($90.19\%$), it maintained balanced precision and recall. In contrast, \textit{Naive Bayes} lagged behind due to its strong independence assumptions. 

Nevertheless, the frozen LSTM embeddings demonstrated exceptional discrimination, resulting in robust performance across various classifier types. The most successful models achieved 100\% or nearly 100\% recall.

\begin{table}[t]
\centering
\caption{Experiment~1 (frozen LSTM): classifier performance on encoder embeddings.}
\label{tab:exp1_frozen_cls}
\begin{tabular}{lcccc}
\toprule
\textbf{Model} & \textbf{Accuracy} & \textbf{Precision} & \textbf{Recall} & \textbf{F1} \\
\midrule
Logistic Regression & 0.93379 & 0.93379 & 1.00000 & 0.96576 \\
Random Forest       & 0.90194 & 0.93668 & 0.95987 & 0.94813 \\
SVM                 & 0.93379 & 0.93379 & 1.00000 & 0.96576 \\
K-Nearest Neighbors & 0.93254 & 0.93371 & 0.99866 & 0.96509 \\
Naive Bayes         & 0.78763 & 0.94050 & 0.82475 & 0.87883 \\
Decision Tree       & 0.88257 & 0.93596 & 0.93846 & 0.93721 \\
Gradient Boosting   & 0.93379 & 0.93433 & 0.99933 & 0.96574 \\
\bottomrule
\end{tabular}
\end{table}

\subsection{Frozen LSTM with Hyperparameter Tuning}

\begin{figure}[t]
  \centering
  \subfloat[Logistic Regression]{%
    \includegraphics[width=0.8\linewidth]{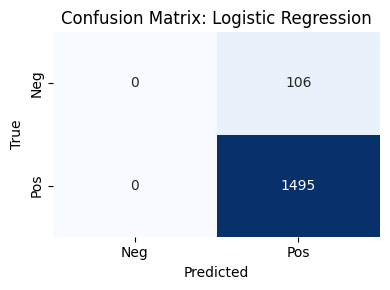}
  }
  \hfill
  \subfloat[Gradient Boosting]{%
    \includegraphics[width=0.8\linewidth]{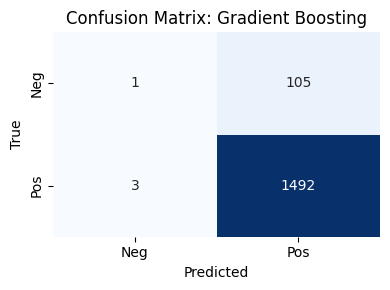}
  }
  \caption{Confusion matrices for tuned classifiers: (a) Logistic Regression and (b) Gradient Boosting.}
  \label{fig:cms}
\end{figure}

\begin{figure}[t]
  \centering
  \includegraphics[width=\linewidth]{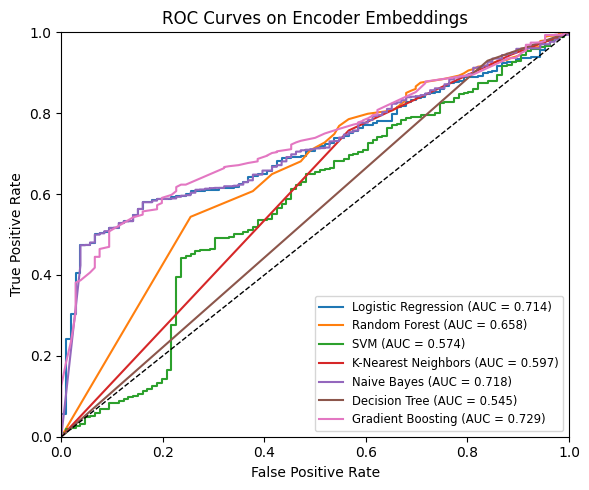}
  \caption{ROC curves on frozen-encoder embeddings (AUC values indicated in legend).}
  \label{fig:roc}
\end{figure}

In this analysis, various optimized classifiers were assessed utilizing embeddings derived from the frozen encoder. Figure~\ref{fig:cms} presents the confusion matrices that encapsulate the classification outcomes for the top two models. For \textit{Logistic Regression}, the findings reveal $\text{TN}=0$, $\text{FP}=106$, $\text{FN}=0$, and $\text{TP}=1495$, signifying perfect recall but a significant number of false positives. In the case of \textit{Gradient Boosting}, the matrix shows $\text{TN}=1$, $\text{FP}=105$, $\text{FN}=3$, and $\text{TP}=1492$, indicating a slight enhancement in specificity with a minor reduction in recall. The ROC--AUC analysis depicted in Figure~\ref{fig:roc} offers further understanding of the models' discriminative capabilities. Gradient Boosting was observed to have attained the highest AUC ($0.729$), with Naive Bayes ($0.718$) and Logistic Regression ($0.714$) following closely. The other classifiers, such as Random Forest ($0.658$), KNN ($0.597$), SVM ($0.574$), and Decision Tree ($0.545$), were found to exhibit relatively lower ability to distinguish between positive and negative classes.

\subsection{Comparative Analysis}

The proposed \ACP framework is compared to existing studies in six critical methodological aspects, as detailed in Table~\ref{tab:comparison_model}:  (i) the integration of the \ACP architecture, (ii) the adoption of an uncertainty-aware PU loss to improve its robustness in positive–unlabeled learning, (iii) the integration of an attention mechanism to improve feature representation, (iv) the incorporation of model validation strategies to ensure reliability, (v) the application of model calibration techniques to improve probability estimation, and (vi) the use of adaptive temperature scaling to refine decision boundaries. 

\begin{table*}[ht]
\centering
\caption{Comparison of Models with Various Techniques}
\label{tab:comparison_model}
\begin{tabular}{|c|c|c|c|c|c|c|}
\hline
\textbf{Study} & 
\makecell{\textbf{\ACP}} & 
\makecell{\textbf{Uncertainty}\\\textbf{Aware PU Loss}} & 
\makecell{\textbf{Attention}\\\textbf{Mechanism}} & 
\makecell{\textbf{Model}\\\textbf{Validation}} & 
\makecell{\textbf{Model}\\\textbf{Calibration}} & 
\makecell{\textbf{Adaptive}\\\textbf{Temperature Scaling}} \\
\hline
Fan et al.~\cite{fan2023self} & \xmark & \xmark & \xmark & \cmark & \xmark & \xmark \\
\hline
Go et al.~\cite{go2020visualization} & \xmark & \xmark & \xmark & \cmark & \xmark & \xmark \\
\hline
Zhang et al.~\cite{zhang2019network} & \xmark & \xmark & \xmark & \cmark & \xmark & \xmark \\
\hline
Zeng et al.~\cite{zeng2019deep} & \xmark & \xmark & \xmark & \cmark & \xmark & \xmark \\
\hline
Nayyar et al.~\cite{nayyar2020recurrent} & \xmark & \xmark & \xmark & \cmark & \xmark & \xmark \\
\hline
Wang et al.~\cite{wang2024pu} & \xmark & \xmark & \xmark & \cmark & \xmark & \xmark \\
\hline
Zhang et al.~\cite{zhang2017poster} & \xmark & \xmark & \xmark & \cmark & \xmark & \xmark \\
\hline
Zhang et al.~\cite{zhang2024semi} & \xmark & \xmark & \xmark & \cmark & \xmark & \xmark \\
\hline
Yang et al.~\cite{ren2014positive} & \xmark & \cmark & \xmark & \cmark & \xmark & \xmark \\
\hline
\rowcolor{green!30}
\ACP & \cmark & \cmark & \cmark & \cmark & \cmark & \cmark \\
\hline
\end{tabular}
\end{table*}

The assessment of previous works (Fan et al.~\cite{fan2023self}, Go et al.~\cite{go2020visualization}, Zhang et al.~\cite{zhang2019network}, Zeng et al.~\cite{zeng2019deep}, Nayyar et al.~\cite{nayyar2020recurrent}, Wang et al.~\cite{wang2024pu}, Zhang et al.~\cite{zhang2024semi}, Yang et al.~\cite{ren2014positive}) indicate that the majority of approaches only employ a subset of these components, typically concentrating on two or three techniques and frequently omitting uncertainty handling, calibration, or adaptive scaling.  In contrast, the proposed \ACP incorporates all six elements into a unified learning pipeline in a unique manner.  Not only does this comprehensive design facilitate the more effective separation of positive and unlabeled instances, but it also enhances the robustness, generalization, and calibration of the system across a variety of datasets.  As a result, \ACP is a substantial methodological improvement over previous methodologies.

\section{Conclusion}
\label{conclusion_futureWork}

This work introduced \ACP, a positive–unlabeled representation learning framework that is uncertainty-aware and incorporates an uncertainty-weighted contrastive loss, self-attention–guided LSTM encoding, and adaptive temperature scaling. The embeddings that were produced by the proposed two-stage training strategy—initial ConPU optimization followed by pseudo-negative triplet refinement—exhibited strong separability, as corroborated by both quantitative metrics and t-SNE visualizations. 

The confusion matrix analysis revealed minimal false negatives (as low as zero) and well-controlled false positives, while the \ACP embeddings consistently enabled high accuracy (up to $93.38\%$), precision above $0.93$, and near-perfect recall across multiple conventional classifiers. The ROC–AUC of Gradient Boosting was the highest ($0.729$), and other top-performing models, including Logistic Regression and SVM, maintained competitive scores. The robustness and generalization capacity of the learned feature space were further validated by the distinct cluster boundaries in both the training and validation splits. 

In contrast to previous PU learning methods, \ACP distinctively incorporates uncertainty modeling, attention mechanisms, calibration, and adaptive scaling into a unified pipeline, thereby addressing the challenges of variable anomaly distributions and noisy supervision. These findings establish \ACP as a scalable and resilient method for high-stakes PU scenarios, with potential applications in fields such as biomedical text mining and cybersecurity threat detection. 

In the future, the possibility of extending the framework to multimodal PU learning will be investigated, with domain adaptation incorporated for cross-dataset generalization and scaling applied to larger real-world datasets with diverse label noise profiles.





\bibliographystyle{IEEEtran}
\bibliography{references}

\end{document}